\documentclass[runningheads]{llncs}

\usepackage [pdftex] {graphicx}
\graphicspath{{../pdf}{../jpeg}}
\DeclareGraphicsExtensions{.pdf,.jpeg,.png}
\usepackage{subfigure}

\usepackage{multicol}  
\usepackage{multirow} 

\usepackage{booktabs}

\usepackage{makecell}

\usepackage{longtable} 

\begin{document}
\title{An Emotion-controlled Dialog Response Generation Model with Dynamic Vocabulary}
%
%
\author{Shuangyong Song \and
Kexin Wang \and
Chao Wang \and
Haiqing Chen \and
Huan Chen
}
\authorrunning{S. Song et al.}
%
\institute{Alibaba Groups, Hangzhou 311121, China\\
\email{\{shuangyong.ssy,marc.wkx,chaowang.wc,haiqing.chenhq,shiwan.ch\}@alibaba-inc.com}}
\maketitle              
\begin{abstract}
In response generation task,
proper sentimental expressions can obviously improve the human-like level of the responses.
However, for real application in
online systems, high QPS (queries per second, an indicator of the flow capacity of on-line systems) is required, and a dynamic vocabulary mechanism has been proved available in improving speed of generative models. In this paper, we proposed an emotion-controlled dialog response generation model based on the dynamic vocabulary mechanism, and the experimental results show the benefit of this model.

\end{abstract}
%
%
%

\section{Introduction}
Chatbots aim to provide users multidimensional human-like services such as shopping guide, chitchat and entertainments. Those kinds of services can significantly help improve users' experience and satisfaction. To create a chatbot capable of communicating with a user at the human level, it is necessary to equip the machine with the ability of perceiving and expressing emotions.

For expressing appropriate emotions in chatbot responses, we linguistically build emotional mappings between user questions and chatbot responses, and then generate emotional responses with an emotion-controlled text generation model.
Compared to retrieval-based chatbots, generation-based chatbots can mostly achieve greater coverage, and generate more proper responses that could have never appeared in the corpus.

However, utilizing response generation model in real online systems has two typical risks: the first one is that the content of generated response is not entirely relevant to the user question, and the second one is that the running speed of response generation models should be further improved to meet the demand of real online systems. In~\cite{2}, a dynamic vocabulary seq2seq (DVS2S) model has been proposed and it can well solve above two risks, since it can especially eliminate abundant noise words from the generation vocabulary, which benefit both the generation speed and the relevance between responses and questions.

In this paper, we try to realize a dynamic vocabulary based emotion-controlled response generation model, which aims to generate emotional responses with high quality and high speed for our chatbot, an industrial intelligent assistant designed for creating an innovative online shopping experience in E-commerce.

\section{Our Model}




\begin{figure}[!h]  
\centering
\includegraphics[width=0.85\textwidth]{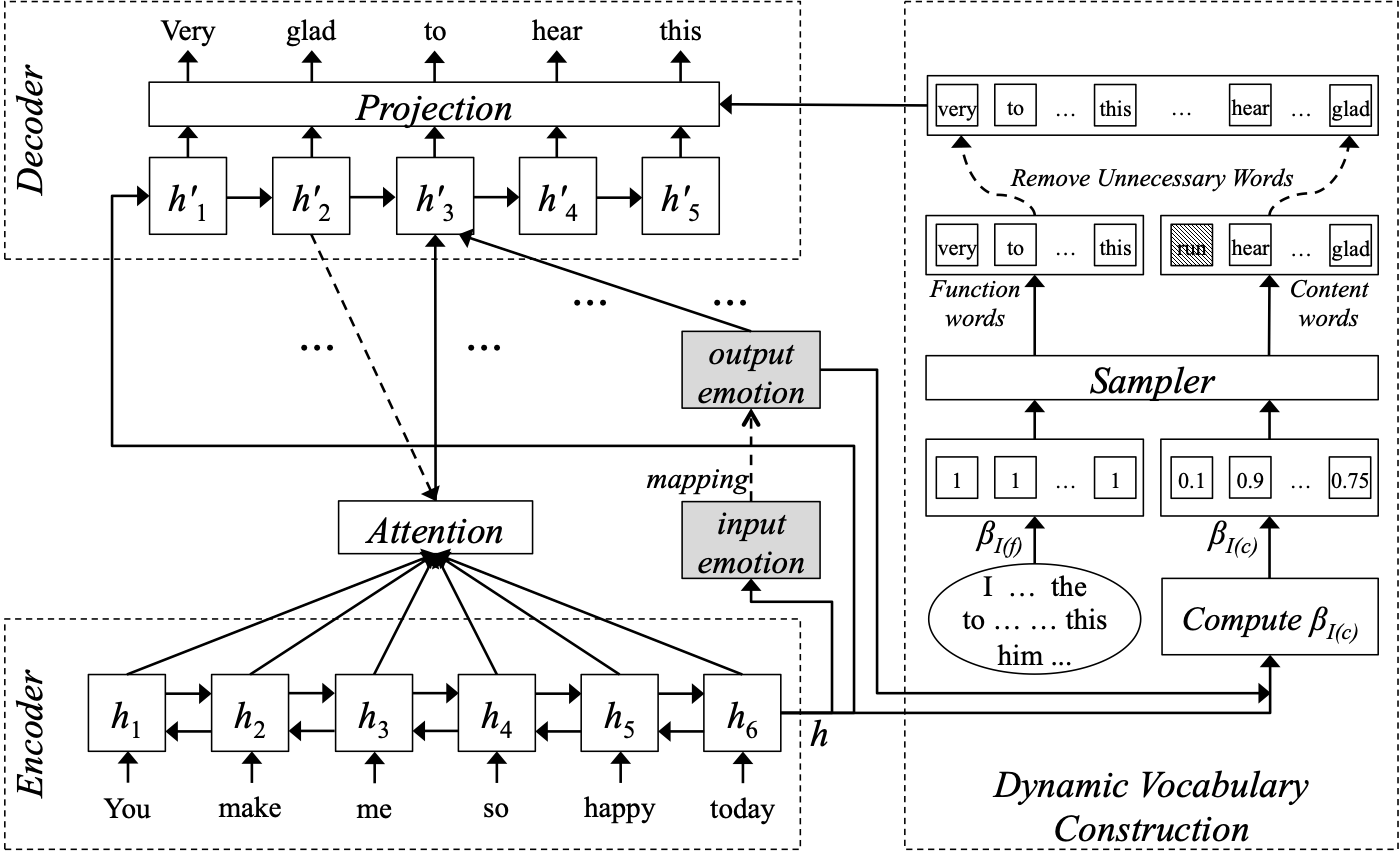}
\caption{Model architecture of DV-ERG.} 
\label{fig1}
\end{figure}


\textbf{1) Emotion Mapping between Questions and Responses}:
We first utilize LEAM (Label-Embedding Attentive Model)~\cite{3} model to realize an emotion classification on user questions, and then we linguistically build emotional mappings between questions and responses, such as a mapping from an ‘abusing’ question to an ‘aggrieved’ or ‘regretful’ response.

\textbf{2) Seq2Seq Model Training}:
For real application in online systems, with high QPS requirement, we just employ typical Bi-GRU (Gated Recurrent Unit) as encoder and GRU~\cite{20} as decoder, with an attention mechanism, instead of very complex models. Employment of typical simple response generation models also shows in other online systems, such as XiaoIce~\cite{4}. The objective function is given below, and compared to a normal function, ours is meanwhile supervised by emotional state e (in formula 1). In our training dataset, all the $\left \langle Question, Response \right \rangle$ pairs comply with the above mapping rules, and the e is set according to the emotion types of responses. In the prediction, e is randomly chosen from candidate mapping emotions of question emotion.


\textbf{3) Vocabulary Model Training}:
Words are separated into function words and content words, and the key is to predict the probability distribution P of content words being selected in the decoder step. We take $\beta _{I(c)}$ as converter from $\left \langle h,e \right \rangle$ to P, where the h is the hidden state of the encoder, and the training task is to optimize $\beta _{I(c)}$.

\textbf{4)Joint Fine-tune}:
We jointly fine-tune the Seq2Seq model and the vocabulary model to further optimize the emotional response generation loss.

\section{Experiments}


\textbf{1) Dataset collection \& Implementation}:
We collect 132,118 frequently asked emotional user questions from the online log of our commercial chatbot, and manually labeled 1 to 3 corresponding emotional responses to each question. Finally we got 308,618 QA pairs as training dataset of the seq2seq model.
In experiments, we use the 300-dimension pre-trained word embeddings, and we set hidden size to 128 both in encoder and decoder, and learning rate to 0.001.

\textbf{2) Baselines}:
We considered the following baselines:
1) \textbf{S2SA}: a standard seq2seq model with an attention mechanism~\cite{5};
2) \textbf{TA-S2S}: the topic-aware seq2seq model proposed in~\cite{6};
3) \textbf{CVAE}: recent work for response generation with a conditional variational auto-encoder~\cite{7};
4) \textbf{DVS2S}: the dynamic vocabulary seq2seq model which allows each input to possess their own vocabulary in decoding~\cite{2}.
With the fine-tune step, we compare 3 different ways:
1) no fine-tune (\textbf{NO-ft});
2) just fine-tune the vocabulary model training step (\textbf{ft-target});
3) fine-tune both Seq2Seq and Vocabulary (\textbf{ft-both}).

\renewcommand{\arraystretch}{1.35} 
\begin{table}[!h]  
  \centering  
  \fontsize{8.5}{8}\selectfont  
  \resizebox{0.8\linewidth}{!}{
    \centering
        \resizebox{\linewidth}{!}{
    		\begin{tabular}{|c|c|c|c|c|c|c|c|c|c|c|}  
    		\hline
   		\multicolumn{2}{|c|}{Models}	& BLEU-2 & Recall & VocSize & Greedy & Average & Extreme & Distinct1 & Distinct2 & si-QPS  	\cr\cline{1-11} 
    		\hline  
    		\hline
    		\multirow{4}{*}{Baselines}		& S2SA 	& 2.79 & 94.21 & 50K & 38.80 & 41.88 & 22.10 & 0.512 & 0.415 & ~68	\cr\cline{2-11} 
								& TAS2S 	& 3.05 & 89.22 & 50K & 38.85 & 43.30 & 34.59 & 0.588 & 0.433 & ~54	\cr\cline{2-11} 
								& CVAE	& 3.02 & 85.55 & 50K & 40.05 & 46.22 & 30.38 & 0.587 & 0.428 & ~35	\cr\cline{2-11} 
    								& DVS2S	& 3.21 & 93.01 & 4K & 42.11 & 49.38 & 37.77 & 0.595 & 0.646 & ~94	\cr\cline{1-11} 
    		\multirow{3}{*}{DV-ERG}		& NO-ft	& 3.38 & 92.16 & 4K & 42.01 & 48.10 & 38.61 & 0.614 & 0.670 & ~92	\cr\cline{2-11} 
								& ft-target	& 3.83 & 92.16 & 4K & 41.79 & 48.15 & 38.50 & 0.605 & 0.655 & ~92	\cr\cline{2-11} 
								& ft-both	& 3.71 & 90.20 & 6.8K & 41.31 & 48.09 & 38.70 & 0.604 & 0.658 & ~92	\cr
    		\hline
    		\end{tabular}  
    }
    }
    \caption{Evaluation results.}  
    \label{tab:table1}  
\end{table}  

\vspace{-0.8cm}

\renewcommand{\arraystretch}{1.35} 
\begin{table}[!h]  
  \centering  
  \fontsize{8.5}{8}\selectfont  
  \resizebox{1.05\linewidth}{!}{
    \centering
        \resizebox{\linewidth}{!}{
    		\begin{tabular}{|c|c|c|p{5cm}|c|}
    		\hline
		Input & Target emotion & Models & Model reply & Labeled reply \cr\cline{1-5} 
    		\hline  
    		\hline
    		\multirow{4}{*}{Then wait patiently!} & \multirow{4}{*}{satisfied}		& Base  & My patience & \multirow{4}{*}{Yeah, well, thank you. You've warmed me up!}	\cr\cline{3-4} 
							&	& NO-ft 	& My patience &	\cr\cline{3-4} 
							&	& ft-target	& Mm-hmm, my little cute~ & 	\cr\cline{3-4} 
    							&	& ft-both	& Ok, mine. &	\cr \hline
    		\multirow{4}{*}{Bullshit!} & \multirow{4}{*}{aggrieved}			& Base  & I don’t know & \multirow{4}{*}{I've done my best.}	\cr\cline{3-4} 
							&	& NO-ft 	& I don’t know &	\cr\cline{3-4} 
							&	& ft-target	& Sorry, I was wrong. &	\cr\cline{3-4} 
    							&	& ft-both	& Oh oh. &	\cr \hline
    		\multirow{4}{*}{Sing a song} & \multirow{4}{*}{non-emotional}		& Base  & Five-star red flag fluttering in the wind, what a resound-ing victory song! & \multirow{4}{*}{No need to know}	\cr\cline{3-4} 
							&	& NO-ft 	& I sing &	\cr\cline{3-4} 
							&	& ft-target	& OK & 	\cr\cline{3-4} 
    							&	& ft-both	& Sing what? &	\cr \hline
    		\end{tabular}  
    }
    }
    \caption{Example results of different varieties of proposed DV-ERG.}  
    \label{tab:table2}  
\end{table}  

\vspace{-0.7cm}

\textbf{3) Metrics}:
In this paper, we refer the evaluation metrics used in~\cite{2}, and further add several metrics:
\textbf{BLEU}:
Since the questions and answers in our chatbot are almost short, we just consider BLEU-2.
\textbf{Recall}:
Only the R@1 metric is relevant in the case of binary classification, which we use here.
\textbf{VocSize}: the size of the vocabulary.
\textbf{si-QPS}: maximum QPS that a single model instance can support.
\textbf{DISTINCT-1} \& \textbf{DISTINCT-2} Following~\cite{2,6}, we employ the Distinct1 and Distinct2 to measure how diverse and informative the generated responses are.
Besides, 3 embedding-based metrics~\cite{21} are used:
\textbf{Greedy},
\textbf{Average}, and
\textbf{Extreme}.

\textbf{4) Experimental results}:
Table 1 gives the evaluation results on different metrics, and we can see ft-target gets the best performance on both BLEU and si-QPS (si-QPS as 92 means about 10.87ms per query). For a real online chatbot, those two metrics are more important than other metrics, so we choose ft-target as the final online response generation model in our chatbot - AliMe.
We qualitatively analyze DV-ERG with some examples from the test data. Table 2 shows several emotional generation results with our models, we can see that most of the generative results are shorter than manually labeled results, this is a common problem of generative models, since short results are ‘safer’ than long sentences in the model training step. However, with emotional empathy and more focused vocabularies, generative results are sometimes even better than manually labeled results. For example,  with a user question as ‘Then wait patiently!’, the manually labeled response is ‘Yeah, well, thank you. You've warmed me up!’. This response is with no problem, but it is better when user get the generated response ‘Mm-hmm, my little cute~’ with ft-target DV-ERG. Another example: with a user question as ‘Sing a song’, the manually labeled response is ‘No need to know’, and this is just a so-so response. This time, the ft-target DV-ERG generate a response as ‘Sing what?’, and this is a more reasonable one.

\section{Conclusion}

In this paper, we proposed an emotion-controlled response generation model based on the dynamic vocabulary mechanism, which can be practically applied to online chat-bots, considering its experimental efficiency and effectiveness. In the future, we will investigate how to apply the emotion analysis technologies and dynamic vocabulary technique to more modules in online chatbots.


%
%
%

\begin{thebibliography}{8}



\bibitem{2}
Wu Y, Wu W, Yang D, Xu C, Li Z. Neural Response Generation With Dynamic Vocabularies. In AAAI 2018, pp. 5594-5601.

\bibitem{3}
Wang G, Li C, Wang W, et al. Joint Embedding of Words and Labels for Text Classification. In ACL 2018, pp. 2321–2331.

\bibitem{4}
Shum H-Y, He X, Li D. From Eliza to XiaoIce: challenges and opportunities with social chatbots. Frontiers of IT \& EE 2019(1): 10-26.

\bibitem{5}
Vinyals, O., and Le, Q. 2015. A neural conversational model. arXiv preprint arXiv:1506.05869.

\bibitem{6}
Xing, C.; Wu, W.; Wu, Y.; Liu, J.; Huang, Y.; Zhou, M.; and Ma, W.-Y. Topic aware neural response generation. In AAAI 2016, pp. 3351-3357.

\bibitem{7}
Zhao, T.; Zhao, R.; and Eskenazi, M. Learning Discourse-level Diversity for Neural Dialog Models using Conditional Variational Autoencoders. ACL 2017, pp. 654-664.













\bibitem{20}
Cho, K., Merrienboer, B. V., Gulcehre, C., Bahdanau, D., Bougares, F., Schwenk, H., Bengio, Y. Learning phrase representations using RNN encoder–decoder for statistical ma-chine translation. In EMNLP 2014, pp. 1724– 1734.

\bibitem{21}
Liu, C-W., Lowe, R., Serban, I., Noseworthy, M., Charlin, L., Pineau, J. How NOT To Evaluate Your Dialogue System: An Empirical Study of Unsupervised Evaluation Metrics for Dialogue Response Generation. In EMNLP 2016, pp. 2122-2132.








\end{thebibliography}
%

\end{document}